%% file: ifacconf.tex
\documentclass{ifacconf}

\usepackage{graphicx}      
\usepackage{natbib}        
\usepackage[acronym, toc]{glossaries}
\usepackage{color}

\usepackage{url}

\usepackage{balance}  
\usepackage{soul}  
\usepackage{amssymb}
\usepackage{tikz,bm,color}
\usepackage{enumerate}
\usepackage{enumitem}
\newcommand{\subscript}[2]{$#1 _ #2$}
\usepackage{rotating}
\usepackage{mathtools, cuted}
\usepackage{booktabs}  
\usepackage{pgfplots}
\usepackage{pgfplotstable}
\usetikzlibrary{external,calc,patterns,decorations.pathmorphing,decorations.markings,decorations.text,automata,arrows,arrows.meta,shapes,positioning, backgrounds,fit,spy}
\usepgfplotslibrary{fillbetween}
\tikzset{weird fill/.style={append after command={
			\pgfextra
			\draw[sharp corners, fill=#1]%
			(\tikzlastnode.west)%
			[rounded corners=3pt] |- (\tikzlastnode.north)%
			[rounded corners=1pt] -| (\tikzlastnode.east)%
			[rounded corners=5pt] |- (\tikzlastnode.south)%
			[rounded corners=0pt] -| (\tikzlastnode.west);
			\endpgfextra}}}

\usepackage[width=0.45\textwidth]{caption}
\usepackage{subcaption}

\definecolor{mygreen}{RGB}{173 221 142}

\pgfplotsset{compat=1.17}  
\begin{document}
\begin{frontmatter}

\input{tex/abbreviations}

\title{Autonomous Navigation in Confined Waters - A COLREGs Rule 9 Compliant Framework\thanksref{footnoteinfo}}

\thanks[footnoteinfo]{This research was sponsored by the Danish Innovation Fund, The Danish Maritime Fund, Orients Fund and the Lauritzen Foundation through the Autonomy part of the ShippingLab project, grant number 8090-00063B. The electronic navigational charts have been provided by the Danish Geodata Agency.
}

\author[First]{Peter N. Hansen} 
\author[First]{Thomas T. Enevoldsen} 
\author[First]{Dimitrios Papageorgiou}
\author[First]{Mogens Blanke}

\address[First]{Technical University of Denmark, Dept. of Electrical and Photonics Engineering, Automation and Control Group, 2800 Kgs. Lyngby, Denmark (e-mail: \{pnha,tthen,dimpa,mb\}@elektro.dtu.dk).}

\input{tex/abstract}


\end{frontmatter}

\input{tex/introduction}
\input{tex/colregs9}

\input{tex/framework}
\input{tex/manoeuvrability_estimation}

\input{tex/results}

\input{tex/conclusion}
\begin{ack}
The authors gratefully acknowledge the Danish Innovation Fund, The Danish Maritime Fund, Orients Fund and the Lauritzen Foundation for support to the ShippingLab Project, Grand Solutions 8090-00063B, which has sponsored this research.
The authors would also like to thank the contributions from Svendborg International Maritime Academy (SIMAC) for their guidance and assistance to the development of the proposed framework.
\end{ack}

\balance
\bibliography{ifacconf}             
\end{document}

%% file: tex/abbreviations.tex
\newacronym{COLREGs}{COLREGs}{International Regulations for Preventing Collisions at Sea}
\newacronym{ECDIS}{ECDIS}{Electronic Chart Display and Information System}
\newacronym{CPA}{CPA}{Closest Point of Approach}
\newacronym{TCPA}{TCPA}{Time for Closest Point of Approach}
\newacronym{DCPA}{DCPA}{Distance from the Closest Point of Approach}
\newacronym{AIS}{AIS}{Automatic Identification System}
\newacronym{GPS}{GPS}{Global Positioning System}
\newacronym{ENC}{ENC}{Electronic Navigational Chart}
\newacronym{INS}{INS}{Inertial Navigation System}
\newacronym{SOG}{SOG}{Speed Over Ground}
\newacronym{COG}{COG}{Course Over Ground}

\newacronym{DFA}{DFA}{Deterministic Finite-State Automata}
\newacronym{DES}{DES}{Discrete-Event Systems}
\newacronym{MPC}{MPC}{Model-Predictive Controller}

\newacronym{COLAV}{COLAV}{Collision Avoidance}
\newacronym{ROS}{ROS}{Rate of Swing}
\newacronym{TV}{TV}{Target Vessel}
\newacronym{OS}{OS}{Own Ship}
\newacronym{ASV}{ASV}{Autonomous Surface Vessel}
\newacronym{SIMAC}{SIMAC}{Svendborg International Maritime Academy}

%% file: tex/abstract.tex
\begin{abstract}                
    Fully or partial autonomous marine vessels are actively being developed by many industry actors.
    In many cases, the autonomous vessels will be operating close to shore, and within range of a Remote Control Center (RCC).
    Close to shore operation requires that the autonomous vessel is able to navigate in close proximity to other autonomous or manned vessels, and possibly in confined waters, while obeying the COLREGs on equal terms as any other vessel at sea.
    In confined waters however, certain COLREGs rules apply, which might alter the expected actions (give-way or stand-on), depending on the manoeuvrability of the vessels.
    This paper presents a Situation Awareness (SAS) framework for autonomous navigation that complies with COLREGs rule 9 (Narrow Channels).
    The proposed solution comprises a method for evaluating the manoeuvrability of a vessel in confined waters, for assessing the applicability of COLREGs rule 9. This feature is then integrated into an already existing SAS framework for facilitating COLREGs-compliant navigation in restricted waters.
    The applicability of the proposed method is demonstrated in simulation using a case study of a small autonomous passenger ferry.
\end{abstract}

%% file: tex/introduction.tex
\section{Introduction}

In order to safely navigate while sailing in confined waters, a navigator on-board any vessel must be able to assess the manoeuvrability of not only ownship, but also that of the surrounding vessels nearby.
To assess another vessels' manoeuvrability, the navigator must take several pieces of information into account, e.g. the size and draught of the vessel, water depth and possible navigational buoys on the seachart, wind speed and direction, current etc.
This information must be combined with an estimate of the intention of the vessel, i.e. its anticipated trajectory.
When a risk of collision occurs, navigators rely on the \gls{COLREGs}, which specifically state the obligations of all vessels that are at a risk of collision.
These obligations however depend on the manoeuvrability of the vessels, and might completely reverse the situation, meaning that a vessel which under normal circumstances has the right of way, may be required to yield.
The specifics of when such obligations shift, are covered by \gls{COLREGs} rule 9.

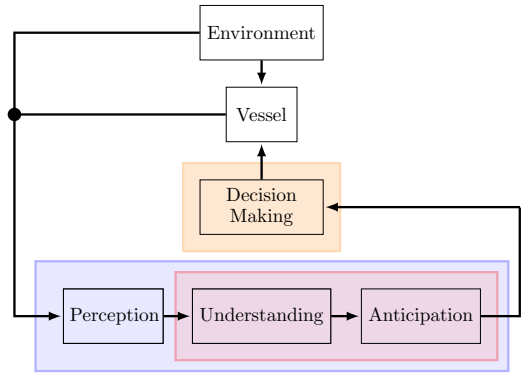
\begin{figure}[bp]
	\begin{center}
		\resizebox{!}{0.55\columnwidth}{
			\tikzsetnextfilename{general_framework}
			\input{tikz/general_framework}
		}
	\end{center}
	\caption{Block diagram of autonomous situation awareness}\label{fig:general_framework}
\end{figure}

Human navigators undergo years of training to be able to safely apply \gls{COLREGs} to avoid risks of collision.
The emergence of maritime  autonomy, either in the form of decision-support systems, or fully autonomous vessels, necessitates the development of \gls{COLREGs}-compliant algorithms that can perform the same collision avoidance tasks.
Such solutions employ modular architectures that comprise sensing, situation evaluation and anticipation and, finally, decision-making as illustrated in Fig.~\ref{fig:general_framework}. Correct interpretation of the situation at sea is essential for safe navigation, especially in confined waters, where the risk of collision is higher given the increased traffic. As such, incorporation of the \gls{COLREGs} rule 9 in the autonomous situation awareness frameworks is of special pertinence.

The development of \gls{COLREGs}-compliant algorithms for situation awareness and collision avoidance have been investigated before, and several frameworks exist in the literature.
The work by \cite{Thyri2020} proposes a collision avoidance algorithm that switches between a set of predefined nominal paths in order to evade traffic, however the proposed method does not evaluate rule 9, as the algorithm tries to avoid all targets, irrespective of the \gls{COLREGs} interpretation.
The use of rule 9 is mentioned in \cite{fang2018simplified} in an overtaking scenario, where the overtaking vessel in a narrow channel should give clear signal to indicate her intention, however the authors do not specify the scenario to an extent that it is possible to evaluate if it is performed in a narrow channel or fairway, nor is the assumption of restricted manoeuvrability mentioned in the scenario description.
The ability for an \gls{ASV} to navigate the confined waters of a river using artificial potential fields is shown in \cite{mei2016COLREGs}.
However the authors do not directly address rule 9, as no assessments of the manoeuvrability of targets or objects is performed.
A deep reinforcement learning based approach is presented in \cite{meyer2020colreg}, in which the authors train an agent to give the right of way to targets that are restricted in their manoeuvrability.
The authors refer to rule 18, which details the obligation of a power driven vessel which must keep out of the way of manoeuvre restricted vessels.
However, in \cite{meyer2020colreg} all target vessels are assumed restricted, and a method for estimating this is not shown.
A framework based on temporal logic is proposed in \cite{krasowski2021temporal} which covers rules beyond the standard collision rules, i.e. rule 6 (safe speed), but does not include an evaluation of rule 9.
One of the most complete (in terms of rules covered) frameworks is presented in \cite{bakdi2022fullest}. 
The authors present a fuzzy logic based framework where the evaluation of rule 9 is based on chart data and the risk of collision with objects and land structures, where the risk of collision is an input to the framework, and is a function of the minimum distances to the nearby objects.
In fact as shown in \cite{burmeister2021autonomous}, most approaches only consider rule 8 and rules 13-17. These particular rules encompass the common collision scenarios, namely overtaking, head-on and crossing scenarios. Where the determination of the stand-on and give-way vessels is not restricted by the environment.
In fact, of the 48 papers discussed in \cite{burmeister2021autonomous} only four papers mention rule 9, while only two of them ``address rule 9" specifically.

While many of the previous studies have shown solutions for navigating in confined waters, such as rivers or other narrow channels, to the best of the authors' knowledge, none of the aforementioned solutions utilize a direct assessment of the target vessels manoeuvrability, something which is directly mentioned in the text of \gls{COLREGs} rule 9.
The work presented in this paper is a framework for the assessment of a target vessels manoeuvrability, and the interpretation and application of \gls{COLREGs} rule 9.
The proposed method is demonstrated through simulation of a real use-case, namely a small harbour ferry crossing the the Aalborg fjord in the northern part of Denmark.

The remainder of the paper is structured as follows: Section \ref{sec:rule9} gives an overview of what rule 9 covers and the challenges it presents.
Section \ref{sec:framework} introduces the proposed framework for automated evaluation and application of rule 9.
Section \ref{sec:manoeuvrability} details the proposed method for estimating manoeuvrability for a given vessel.
Section \ref{sec:results} discusses the simulation scenarios and results.
Finally, Section \ref{sec:conclusion} provides concluding remarks and comments on future work.

%% file: tikz/general_framework.tex
\begin{tikzpicture}[shorten >=1pt,on grid,initial distance=1cm,
  every initial by arrow/.style={*-latex}]

	\node[minimum height=1.0cm,minimum width=1.1cm, text width=3em, align=center, draw](vessel) {Vessel};
	\node[minimum height=1.0cm,minimum width=1.1cm, text width=5.7124em, align=center, above of=vessel, node distance = 1.5cm, draw](environment) {Environment};
	\node[minimum height=1.0cm,minimum width=1.1cm, text width=5.7124em, align=center, below of=vessel, node distance = 1.7cm](decision) {};

	\node[minimum height=1.0cm,minimum width=1.1cm, text width=4.5em, align=center, below of=decision, node distance = 2cm, xshift=-2.7cm, draw](perception) {};
	\node[minimum height=1.0cm,minimum width=1.1cm, text width=6.5em, align=center, below of=decision, node distance = 2cm, draw] (understanding) {};
	\node[minimum height=1.0cm,minimum width=1.1cm, text width=5.5em, align=center, below of=decision, node distance = 2cm, xshift=2.9cm] (anticipation) {};
	
	\node[left of=vessel, node distance=4.5cm](dummyleft){};
	\node[right of=decision, node distance=4.7cm](dummyright){};
	
	\fill (dummyleft) circle [radius=3.5pt];
	
	\draw[blue,very thick, fill=blue!30, opacity=0.3] ($(perception.north west)+(-0.5,0.5)$)  rectangle ($(anticipation.south east)+(0.5,-0.5)$);
	\draw[red,very thick, fill=red!30, opacity=0.3] ($(understanding.north west)+(-0.3,0.3)$)  rectangle ($(anticipation.south east)+(0.3,-0.3)$);
	\draw[orange,very thick, fill=orange!60, opacity=0.3] ($(decision.north west)+(-0.3,0.3)$)  rectangle ($(decision.south east)+(0.3,-0.3)$);

	\draw[-, very thick] (environment.west) -| (dummyleft.south);
	\draw[-latex, very thick] (environment.south) -| (vessel.north);
	\draw[-, very thick] (vessel.west) -- (dummyleft.west);
	\draw[-latex, very thick] (dummyleft.north) |- (perception.west);
	\draw[-latex, very thick] (perception.east) |- (understanding.west);
	\draw[-latex, very thick] (understanding.east) |- (anticipation.west);
	\draw[-, very thick] (anticipation.east) -| ([yshift=1.5mm] dummyright.south);
	\draw[-latex, very thick] ([xshift=1.3mm]dummyright.west) -- (decision.east);
	\draw[-latex, very thick] (decision.north) -- (vessel.south);

	\node[minimum height=1.0cm,minimum width=1.1cm, text width=4.5em, align=center, draw] at (perception) {Perception};
	\node[minimum height=1.0cm,minimum width=1.1cm, text width=6.5em, align=center, draw] at (understanding) {Understanding};
	\node[minimum height=1.0cm,minimum width=1.1cm, text width=5.5em, align=center, draw] at (anticipation) {Anticipation};
	\node[minimum height=1.0cm,minimum width=1.1cm, text width=5.7124em, align=center, draw] at (decision) {Decision Making};
\end{tikzpicture}

%% file: tex/colregs9.tex
\section{Overview of COLREGs rule 9}\label{sec:rule9}  

As previously mentioned the majority of the proposed frameworks in the literature only handle (COLREGs) rules 8 \& 13-17, which dictates the behaviours of vessels when a collision risk is present.
However, rules 13-17 does not take into account any environmental or vessel related factors, which might impose restrictions on the manoeuvrability on any of the vessels in the situation.
Factors that restrict manoeuvrability include, handling capabilities of the vessel (e.g. stopping distance, turn rate, etc.) and draught of the vessel (decreasing the amount of safe water).
In fact rule 3 section h) in \citep{COLREGS} defines a ``vessel constrained by her draught" as: 
\begin{quote}
    \textit{A power-driven vessel which, because of her draught in relation to available depth and width of navigable water, is severely restricted in her ability to deviate from the course she is following.}
\end{quote}
The above definition seems obvious, but it does not define any metric for evaluating when a vessel is ``severely restricted", this is up to the navigator to assess based on navigational experience and the given situation.
A subset of the paragraphs of \gls{COLREGs} rule 9 are detailed below (for the full description see \cite{COLREGS}).
\begin{description}
    \item[a)] \textit{A vessel proceeding along the course of a narrow channel or fairway shall keep as near to the outer limit or the channel or fairway which lies on her starboard side as is safe and practicable.}
    
    \item[b)] \textit{A vessel of less than 20 m in length or a sailing vessel shall not impede the passage of a vessel which can safely navigate only within a narrow channel or fairway.}
    
    \item[d)]  \textit{A vessel shall not cross a narrow channel or fairway if such crossing impedes the passage of a vessel which can safely navigate only within such channel or fairway. The latter vessel may use the sound signal prescribed in rule 34 (d) if in doubt as to the intention of the crossing vessel.}

\end{description}
It is clear from the description above that navigators are expected to maintain safety for ownship, and not to impede the safety of other vessels.
The description mentions ``narrow channel or fairway" as a requirement for the applicability of rule 9, however this is not easily defined as mentioned in \cite{cockcroft2003guide}.
The definition of the terms fairway and narrow channel are according to \cite{cockcroft2003guide}, and paraphrased as follows:
\begin{description}
    \item[Fairway] A navigable passage of water, or dredged channel maintained by the port authority marked by pecked lines on the chart, i.e. a deep water channel.
    \item[Narrow channel] The navigable width of water between lines of buoys.
\end{description}
The above definition is useful for distinguishing between the two terms, but it does not provide any tangible measure for which circumstances something is a narrow channel.
In \cite{cockcroft2003guide} it is noted that \textit{``seamen usually navigate the locality and the advice given by the Elder Brethren"}, indicating that the definition of narrow passage is based on the common consensus and experience of the local seamen.
In fact, passages approximately 2 nautical miles wide have in some cases been considered "narrow" \citep{cockcroft2003guide}.

Then how can one distinguish if a channel is narrow? 
Looking closer section d) in rule 9 \citep{COLREGS}, it is evident that the navigator must estimate if a vessel ``can safely navigate only within such channel or fairway", juxtaposing this with the definition in rule 3, where a vessel is ``restricted in her ability to deviate from the course she is following", we arrive at the following: The inability to deviate from a current course imposes a restriction on a vessel, and if such restrictions are due to environmental constraints, e.g. deep water channel, shallow water, etc., then a vessel can be considered restricted.

%% file: tex/framework.tex
\section{Design of Framework}\label{sec:framework}

For an autonomous system to interpret the applicability of \gls{COLREGs} rule 9 and be compliant, it is required that the system is able to assess the manoeuvrability restrictions on any vessel.
With the introduction of \gls{AIS}, vessels can broadcast information regarding themselves, such as, length, width and navigational status.
The navigational status field can be used to broadcast \emph{restricted manoeuvrability} and \emph{draught constrained} to other vessels, indicating the inability to deviate from it's current path, e.g. a container vessel approaching port in a dredged channel or a fishing vessel currently fishing, etc. 
It should be noted that a given vessel also can report restricted manoeuvrability using flags and lights (rule 28).
However, not all vessels are equipped with \gls{AIS} transponders, and even worse, not all vessels update and broadcast the "correct" status.
As it cannot be assumed that the information is available from \gls{AIS}, an autonomous system would need additional information in order to estimate the manoeuvrability of other vessels.

The autonomous system is equipped with a wide variety of sensors such as radar, GPS, \gls{ECDIS} and including an array of cameras for visual detection.
All this sensory information is processed by a fusion algorithm, which outputs a set of detected targets and their attributes.
Each target detected by the autonomous system has the following attributes:
\begin{itemize}
    \item Position
    \item Speed
    \item Course
    \item Dimensional data: length, width \& draught (either from AIS or estimated by visual systems)
    \item Navigational status (only for targets with \gls{AIS} transponders)
    \item Type (vessel, buoy, kayak, etc.)
    \item Sub-type (type of vessel, type of buoy, etc.)
\end{itemize}
A complete framework for evaluating current and future situation at sea was introduced in \citep{Papageorgiou2019} and extended in \citep{hansen2020colregs}. This situation awareness solution is using tools from the arsenal of \gls{DES} theory and specifically, deterministic automata \citep{cassandras2008a} for determining whether a vessel (\gls{OS} or \gls{TV}) will stand on or give way based on sensory information about its type, position, speed etc.

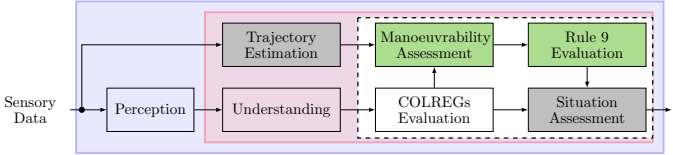
\begin{figure}
    \begin{center}
		\resizebox{!}{0.23\columnwidth}{
			\tikzsetnextfilename{automaton_D}
			\input{tikz/extended_framework1}
		}
	\end{center}
	\caption{Expanded view of how the proposed framework fit into the existing work}
	\label{fig:extended_framework}
\end{figure}

This study extends the framework proposed in \citep{hansen2020colregs} by introducing a new automaton that assesses whether \gls{COLREGs} rule 9 should be considered or not when evaluating the situation at sea. This assessment automaton $G_d$, is defined as the five-tuple
\begin{equation}\label{eq:assessment_automaton_def}
    G_d \triangleq (\mathcal{D}, E_d, f_d, D_1,\{D_1, D_4\}) ,
\end{equation}
where the set of marked states contain the states $D_1$, $D_4$, as they represent the completion of a manoeuvrability assessment.
\begin{figure}[t]
	\begin{center}
		\resizebox{!}{0.55\columnwidth}{
			\tikzsetnextfilename{automaton_A}
			\input{tikz/automaton_A}
		}
	\end{center}
	\caption{State transition diagram of the COLREGs rule 9 assessment automaton $G_d$.}\label{fig:automaton_E}
\end{figure}
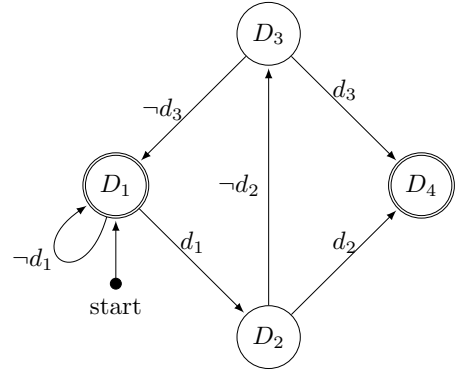
The state transition diagram of the assessment automaton is shown in Fig.~\ref{fig:automaton_E}.
The discrete states $D_i:\bigcup\limits_{i = 1}^{4}\{D_i\}\triangleq\mathcal{D}$ concerning the assessment of detected objects are listed below:
\begin{enumerate}[label=\subscript{D}{{\arabic*}}:]
	\item Check if \gls{TV} needs to give way
	\item Check if \gls{TV} is reporting "restricted manoeuvrability"
	\item Evaluate manoeuvrability of \gls{TV}.
	\item Apply rule 9: inversion of Give-way/Stand-on.
\end{enumerate}
Since the set of marked states $\{D_1, D_4\}$ are \emph{coaccesible} from all non-marked states, the assessment automaton is non-blocking \citep{cassandras2008a}.
The event set associated to the discrete states set $\mathcal{D}$ is defined as $E_d \triangleq \bigcup\limits_{i = 1}^{3}\{d_i,\lnot d_i\}$ where the event set $d_i$ is generated by the assessment automaton $G_d$.
\begin{enumerate}[label=\subscript{d}{{\arabic*}}:]
	\item \label{event:a_target_give_way} \gls{TV} needs to give way.
	\item \label{event:a_target_ais_restricted} \gls{TV} is reporting \textit{restricted manoeuvrability}.
	\item \label{event:a_target_estimate_restricted} \gls{TV} is estimated to have \textit{restricted manoeuvrability}.
\end{enumerate}
The overall action flow of the extended situation assessment framework is illustrated in Fig. \ref{fig:extended_framework}. 
A \emph{perception} module comprising sensor fusion functionalities processes measured data (camera images, \gls{ECDIS}, radar, GPS, etc.) and delivers information about the position, bearing, speed, type, anticipated future trajectory etc. of every vessel that is of interest for the assessment of collision risk. 
In case, such risk exists, application of the \gls{COLREGs} suggests whether a vessel needs to give way or not. 
The latter is the output of the automaton $G_a$ defined in \citep{hansen2020colregs}. 
This triggers the events $\{d_1, \lnot d_1\}$ of $G_d$, which lead to the assessment of the vessel's manoeuvrability and eventually to deciding whether rule 9 is applicable.
Finally, the situation is re-evaluated and sent to the decision logic of the autonomous system.

%% file: tikz/automaton_A.tex
\begin{tikzpicture}[shorten >=1pt,on grid,initial distance=1cm,
  every initial by arrow/.style={*-latex}]
	\node[state,accepting,initial below](D_1) at (0,0) {$D_1$};
  	\node[state,node distance=3cm] (D_2) [below right=of D_1] {$D_2$};
  	\node[state,node distance=3cm] (D_3) [above right=of D_1] {$D_3$};
  	\node[state,accepting,node distance=3cm] (D_4) [above right=of D_2] {$D_4$};

    \path[-latex,every loop/.style={min distance=10mm,in=215,out=254,looseness=10}] (D_1) edge node [above] {$d_1$} (D_2)
        edge [loop left] node {$\lnot d_1$} ();
    \path[-latex] (D_2) edge node [above] {$d_2$} (D_4)
        (D_2) edge node [left] {$\lnot d_2$} (D_3);
    \path[-latex] (D_3) edge [] node [above] {$d_3$} (D_4)
        (D_3) edge [] node [left] {$\lnot d_3$} (D_1);
\end{tikzpicture}

%% file: tikz/extended_framework1.tex
\begin{tikzpicture}
	\draw[blue,very thick, fill=blue!30, opacity=0.3] ($(-1.7, 2.5)$)  rectangle ($(11.75, -1)$);
	\draw[red,very thick, fill=red!30, opacity=0.3] ($(1.25, 2.25)$)  rectangle ($(11.5, -0.75)$);
	\draw[black, thick, dashed, fill=white, opacity=1] ($(4.75, 2.125)$)  rectangle ($(11.5, -0.65)$);
	\draw node[minimum height=1.0cm,minimum width=1.1cm, text width=5em, align=center, draw] (per) {Perception};
	\node[left of = per, node distance = 2.75 cm, minimum height=1.0cm,minimum width=1.1cm, text width=4.5em, align=center](dat){Sensory Data};
	\draw[-latex](dat.east)--(per.west);
	\fill([xshift = 7.5pt]dat.east) circle [radius=2pt];
	\draw node[right of = per, minimum height=1.0cm,minimum width=1.1cm, text width=7em, align=center, node distance = 3 cm, draw] (und) {Understanding};
	\draw node[above of = und, minimum height=1.0cm,minimum width=1.1cm, text width=7em, align=center, node distance = 1.5 cm, fill=lightgray, draw] (traES) {Trajectory Estimation};
	\draw[-latex](per.east)--(und.west);
	\draw[-latex]([xshift=7.5pt]dat.east)|-(traES.west);
	\draw node[right of = und, minimum height=1.0cm,minimum width=1.1cm, text width=7em, align=center, node distance = 3.5 cm, fill=white, draw] (COL) {COLREGs Evaluation};
	\draw node[above of = COL, minimum height=1.0cm,minimum width=1.1cm, text width=7em, align=center, node distance = 1.5 cm, fill=mygreen, draw] (ManoE) {Manoeuvrability Assessment};
	\draw[-latex](und.east)--(COL.west);
	\draw[-latex](traES.east)--(ManoE.west);
	\draw node[right of = COL, minimum height=1.0cm,minimum width=1.1cm, text width=7em, align=center, node distance = 3.5 cm, fill=lightgray, draw] (SAS) {Situation Assessment};
	\draw node[above of = SAS, minimum height=1.0cm,minimum width=1.1cm, text width=7em, align=center, node distance = 1.5 cm, fill=mygreen, draw] (RulEninE) {Rule 9 Evaluation};
	\node[right of = SAS, node distance = 2.5 cm, minimum height=1.0cm,minimum width=1.1cm, text width=2.5em, align=center](out){};
	\draw[-latex](COL.east)--(SAS.west);
	\draw[-latex](COL.north)--(ManoE.south);
	\draw[-latex](ManoE.east)--(RulEninE.west);
	\draw[-latex](RulEninE.south)--(SAS.north);
	\draw[-latex](SAS.east)--(out.west);
\end{tikzpicture}

%% file: tex/manoeuvrability_estimation.tex
\section{Assessing  vessel manoeuvrability}\label{sec:manoeuvrability}
Evaluating whether or not a given vessel is restricted by its manoeuvrability boils down to accessing whether not a given vessel can adequately perform an evasive maneuver, if rules 13-15 apply. 
By computing the maneuvering area of a given vessel based on a metric or a subset of its actions, one can to a certain extent determine if the vessel is capable of giving way.
Such an approach is demonstrated by \cite{wang2020autonomous}, where a description of the maneuvering (reachable) area is computed based on the Nomoto model, which is then evaluated to determine the risk of a given situation.
By evaluating the maneuvering area, within the immediate future, one can determine whether or not subset of the maneuvers may cause unsafe or potentially dangerous situations. 
The following section details a proposed scheme of computing and accessing the maneuvering options available to a given vessel in confined waters.

\subsection{Manoeuvrability Estimation}

\begin{figure}[tb]
	\begin{center}
		\resizebox{!}{0.75\columnwidth}{
			\tikzsetnextfilename{rate_of_swing_plot}
			\input{tikz/rate_of_swing_plot}
		}
	\end{center}
	\caption{Shallow water rate of swing for the 14 different vessels.}\label{fig:rate_of_swing_plot}
\end{figure}
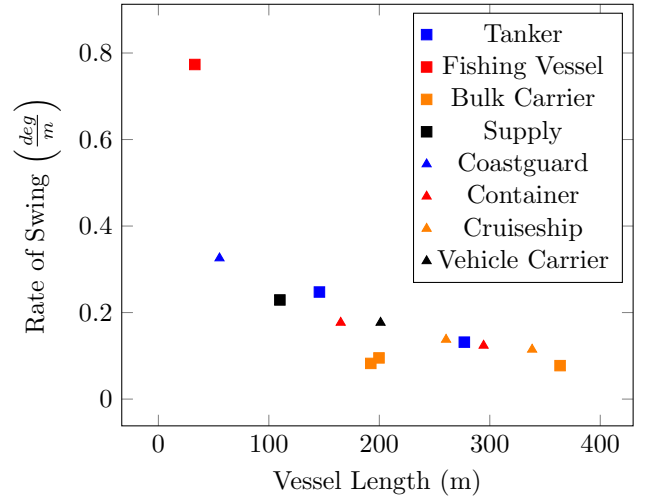

\begin{figure}[tb]
    \centering
    \begin{subfigure}[b]{0.49\textwidth}
        \centering
        \input{tikz/rate_of_swing_projection}
        \caption{Projection of the calculated $p'$ based on the  rate of swing $\delta'_{swing}$.}
        \label{fig:rate_of_swing_projection}
    \end{subfigure}
    \hfill
    \begin{subfigure}[b]{0.49\textwidth}
        \centering
        \input{tikz/rate_of_swing_position}
        \caption{Future relative position $p'_{body}$ of a starboard turn, given heading change $\psi'$ and distance traveled along the arc length $l$.}
        \label{fig:rate_of_swing_position}
    \end{subfigure}
    \caption{Projection of the future position of a target given $\delta'_{swing}$ }
\end{figure}
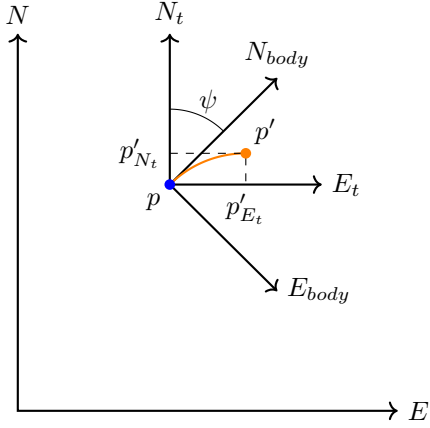
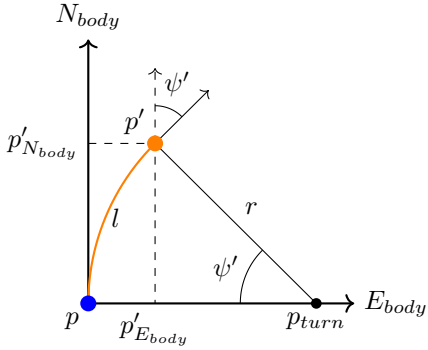

When estimating a vessels manoeuvrability, many factors must be taken into account; size (length and width), thruster and rudder configuration (multiple thrusters and rudders, azimuth, etc.), loading conditions (for container, tanker and bulk carriers), just to name a few.
However, the before mentioned quantities can be difficult to obtain or estimate, if not impossible, even for a human navigator.
As such the autonomous system must be able to estimate of the manoeuvrability of other vessels, based on information available to the system.
One way to estimate the manoeuvrability of a vessel based on the dimensions (i.e. length), is to compare against the turning circles of vessels of similar dimensions.
The definitions of the turning circle are given in \cite{IMO_manoeuvrabilty}.
Table \ref{tab:rate_of_swing_table} contains the length $l$, breadth $b$, rudder type, max rudder angle $\delta_{max}$ and turning circle radius $r_{turn}$ of 14 different vessels used in simulation by experts at \gls{SIMAC} during the training and evaluation of master mariners.
\begin{table}[tb]
    \centering
    \caption{Overview of vessels used for calculation of rate of swing}
    \label{tab:rate_of_swing_table}
    \resizebox{\columnwidth}{!}{
        \begin{tabular}{ l r r l r r }            
            \toprule
            Vessel type     & $l$ (m)    & $b$ (m)   & rudder & $\delta_{max}$ (deg) & $r_{turn}$ (m) \\
            \midrule
            fishing & 33.0 & 8.0 & normal & 35.0 & 74.1 \\
            bulk carrier & 192.3 & 20.4 & normal & 35.0 & 694.5 \\
            bulk carrier & 199.7 & 31.8 & normal & 44.0 & 601.9 \\
            bulk carrier & 363.6 & 65.0 & normal & 35.0 & 740.8 \\
            unknown & 265.6 & 44.0 & becker & 65.0 & 342.6 \\
            supply & 110.0 & 24.0 & normal & 35.0 & 250.0 \\
            coastguard & 55.3 & 9.7 & normal & 35.0 & 175.9 \\
            container & 294.4 & 32.4 & normal & 37.0 & 463.0 \\
            container & 165.0 & 27.5 & normal & 35.0 & 324.1 \\
            cruiseship & 338.3 & 56.0 & azipod & 180.0 & 500.0 \\
            cruiseship & 260.4 & 37.4 & normal & 42.0 & 416.7 \\
            vehicle carrier & 201.1 & 34.1 & normal & 35.0 & 324.1 \\
            tanker & 145.8 & 24.0 & becker & 60.0 & 231.5 \\
            tanker & 277.0 & 42.7 & normal & 45.0 & 435.2 \\
            \bottomrule
        \end{tabular}
    }
\end{table}
Based on the turning circles of the vessel, it is possible to calculate the heading change per meter travelled along the turning circle, called \emph{rate of swing} denoted as $\delta_{swing}$
\begin{equation}\label{eq:rate_of_swing}
    \delta_{swing} = \frac{360}{c} = \frac{360}{2 \pi r_{turn}} \quad \left(\frac{deg}{m}\right).
\end{equation}
The $\delta_{swing}$ described in \eqref{eq:rate_of_swing} is calculated for all the vessels in Table \ref{tab:rate_of_swing_table}, with the resulting values plotted in Fig.~\ref{fig:rate_of_swing_plot}.
While there is no clear linear trend in the $\delta_{swing}$ values (unless the fishing vessel is excluded), Fig.~ \ref{fig:rate_of_swing_plot} does give a lower bound for $\delta_{swing}$ which is $\approx 0.1 (deg/m)$.
However, since the turning circles are defined as the circle a vessel can make by turning its rudder by 35 degrees (or the maximum angle if less than 35) \citep{IMO_manoeuvrabilty}, the values in Fig.~\ref{fig:rate_of_swing_plot} illustrates the maximum $\delta_{swing}$ vessels can perform, something that for the majority of the vessels, is highly undesirable in all but a very few situations.
To represent $\delta_{swing}$ in a more realistic region, the value is scaled by a linear parameter $\alpha$. 
\begin{equation}\label{eq:rate_of_swing_scaled}
    \delta'_{swing} = \alpha \delta_{swing},
\end{equation}
where $\alpha \in [0, 1]$. 
This assumes the turning radius scales linearly with the rudder angle, this relation is certainly way to simplistic in nature, although the relationship in \eqref{eq:rate_of_swing_scaled} could easily be replaced with a non-linear relationship, but this is outside the scope of this paper.
The future position of a vessel $p'$, given a position $p$, heading $\psi$, rate of swing $\delta'_{swing}$, is given as
\begin{align}\label{eq:pprime}
    p' =  
    \begin{bmatrix}
        p'_N \\
        p'_E
    \end{bmatrix}
    &= R(\psi) p'_{body} + p,
\end{align}
where $\psi$ is the current compass course of the vessel, $R(\psi)$ is the rotation matrix that projects a point in body-frame to NED-frame, and $p'_{body}$ is the relative point in body-frame defined as
\begin{align}\label{eq:p'_body}
    p'_{body} &=
    \begin{bmatrix}
        p'_N \\
        p'_E
    \end{bmatrix}_{body}
    = 
    r
    \begin{bmatrix}
        \sin(\psi') \\
        ( 1 - \cos(\psi'))
    \end{bmatrix},
\end{align}
where $r$ is the turning radius and $\psi'$ is the heading change.
The trigonometric relations are illustrated in Fig.~\ref{fig:rate_of_swing_projection} and \ref{fig:rate_of_swing_position}.
The heading change $\psi'$ can be expressed in terms of $\delta'_{swing}$
\begin{equation}\label{eq:psi'}
    \psi' = l \delta'_{swing},
\end{equation}
where $l$ is the distance travelled along the circle with radius $r$, which can be expressed as
\begin{align}\label{eq:r}
    r &= \frac{l}{\psi'} = \frac{l}{l \delta'_{swing}} = \frac{1}{\delta'_{swing}}.
\end{align}
Inserting \eqref{eq:psi'} and \eqref{eq:r} into \eqref{eq:p'_body} yields
\begin{align}
    p'_{body} &=
    \frac{1}{\delta'_{swing}}
    \begin{bmatrix}
        \sin(l \delta'_{swing}) \\
        1 - \cos(l \delta'_{swing})
    \end{bmatrix}.
\end{align}
One can then estimate the clockwise $p'_{\text{CW}}$ and counterclockwise $p'_{\text{CCW}}$ position estimate, based whether $\psi'$ is positive or negative respectively.

\subsection{Manoeuvrability Evaluation}
\begin{figure}[t]
        \centering
        \includegraphics[width=\columnwidth]{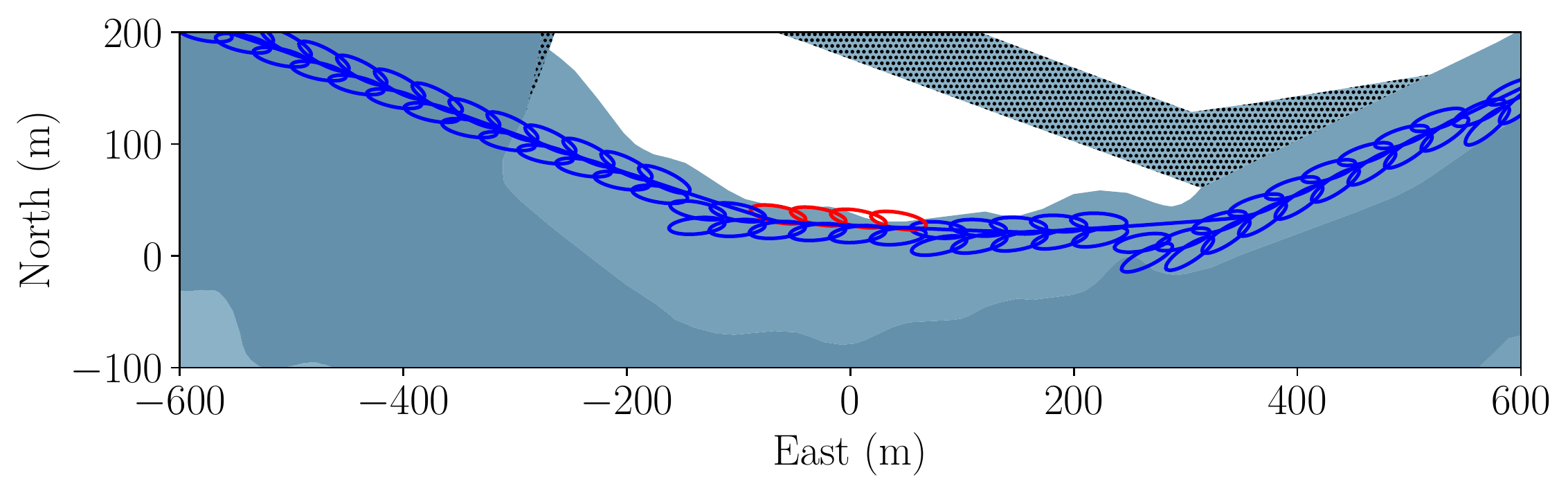}
        \caption{Accessing the manoeuvrability of a target vessel using $\delta'_{swing}$ and ship domain. Red ellipses indicate areas where the vessel is restricted in its manoeuvrability, and blue indicate areas where the vessel is not restricted.}
        \label{fig:ros_shipdomain}
\end{figure}
Given an estimated trajectory of a given vessel, as well as the corresponding estimates of $p'_{\text{CW}}$ and $p'_{\text{CCW}}$ for each time step, one can conservatively evaluate whether or not a given vessel may ground. At each of the estimated positions $p'$, the ship domain can be checked with respect to feasible depth contours.

Ship domains (comfort zones) are a useful tool to maintain an adequate safety distance from other vessels and the environment.
\cite{hansen2013empirical} detailed an elliptical description based on AIS data, scaled by the vessels ship length. \cite{pietrzykowski2021effective} used a training simulator to estimate elliptical ship domains, dependant on ship size and speed. \cite{wang2016empirically} proposed a polygonal domain that iteratively updates with ship speed.
In \cite{du2021empirical} the authors provide an overview of additional ship domain estimation techniques, as well as proposing a method for computing the ship domain based on the available maneuvering margin.

To demonstrate the manoeuvrability estimator, an elliptical comfort zone is placed at each $p'$, where the comfort zone is described by
\begin{equation} 
\begin{split}
    \frac{\left(p'_E(t) \sin \psi(t) + p'_N(t) \cos \psi(t)\right)^2}{a^2} +\\ \frac{\left(p'_E(t) \cos \psi(t) - p'_E \sin \psi(t) \right)^2}{b^2}\leq  1,
\end{split}
\end{equation}
where $p'_N(t)$, $p'_E(t)$ and $\psi(t)$ is the target vessels north-east position and heading in time.
If one of the ellipses intersects with the boundary of the feasible depth contour, the vessel is deemed to have restricted manoeuvrability at that point in time. 
Fig.~\ref{fig:ros_shipdomain} demonstrates the intersection between the elliptical ship domains at each time step, where the red ellipses violate the restrictions posed by the feasible water depths.

The proposed manoeuvrability estimator can be evaluated using any comfort zone of choice. 
For a less conservative and more accurate estimate of the manoeuvrability, one should use a domain that takes into account the characteristics of the given environment and ship. 
These characteristics can be obtained through studies such as those presented above, as well as through expert knowledge.
By leveraging expert knowledge of the local confined and inner coastal waters, one can apply vessel specific restrictions.

%% file: tikz/rate_of_swing_plot.tex
\pgfplotstableread[col sep = comma]{data/rate_of_swing.csv}\mytable

\begin{tikzpicture}

\begin{axis}[
    enlargelimits=0.2,
    xlabel=Vessel Length (m),
    ylabel=Rate of Swing $\left(\frac{deg}{m}\right)$
]

\addplot[
    scatter/classes={
        tanker={mark=square*,blue},
        fishing={mark=square*,red},
        bulk_carrier={mark=square*,orange},
        supply={mark=square*,black},
        coastguard={mark=triangle*,blue},
        container={mark=triangle*,red},
        cruiseship={mark=triangle*,orange}, 
        vehicle_carrier={mark=triangle*,black}
    },
    scatter, 
    only marks, 
    scatter src=explicit symbolic] 
    table [%
        x = length, 
        y = shallow_water_swing_rate,
        meta = class,
        col sep = comma
    ]{\mytable};
    \legend{
        Tanker,
        Fishing Vessel,
        Bulk Carrier,
        Supply,
        Coastguard,
        Container,
        Cruiseship, 
        Vehicle Carrier,
    }

\end{axis}
\end{tikzpicture}

%% file: tikz/rate_of_swing_projection.tex
\begin{tikzpicture}

    \draw [<->,thick] (0,5) node (yaxis) [above] {$N$}
        |- (5,0) node (xaxis) [right] {$E$};
    
    \coordinate (p) at (2, 3);
    
    \coordinate (p') at (3, {sqrt(2)+2});
    
    \draw [<->,thick, rotate=-45] ($(0,2) + (p)$) node (yaxis) [above] {$N_{body}$}
        |- ($(2,0) + (p)$) node (xaxis) [right] {$E_{body}$};
    
    \draw [<->,thick] ($(0,2) + (p)$) node (y_trans) [above] {$N_t$}
        |- ($(2,0) + (p)$) node (x_trans) [right] {$E_t$};
    
    \draw [thin,domain=45:90] plot ({cos(\x) + 2}, {sin(\x) + 3}) node [above right, xshift=7.5pt, yshift=-5pt] {$\psi$};
    
    \draw [orange,thick,domain=90:135] plot ({sqrt(2)*cos(\x) + 3}, {sqrt(2)*sin(\x)+2});
    
    \draw[dashed] (y_trans |- p') node[left] {$p'_{N_t}$} -| (x_trans -| p') node[below] {$p'_{E_t}$};

    \fill[orange] (p') circle (2pt) node [above right, color=black] {$p'$};
    \fill[blue] (p) circle (2pt) node [below left, color=black] {$p$};
\end{tikzpicture}

%% file: tikz/rate_of_swing_position.tex
\begin{tikzpicture}
    
    \draw [<->,thick] (0,3.5) node (yaxis) [above] {$N_{body}$}
        |- (3.5,0) node (xaxis) [right] {$E_{body}$};
        
    \coordinate (p) at (0, 0);
    
    \draw [orange,thick,domain=135:180] plot ({3*cos(\x)+3}, {3*sin(\x)}) node [ above right,color=black, xshift=5pt, yshift=25pt] {$l$};
    \coordinate (p') at ({3*cos(135)+3}, {3*sin(135)});
    \draw[->] ({3*cos(135)+3}, {3*sin(135)}) -- ({3*cos(135)+3 + cos(45)}, {3*sin(135) + sin(45)});
    \draw[->, dashed] ({3*cos(135)+3}, {3*sin(135)}) -- ({3*cos(135)+3}, {3*sin(135) + 1});
    \draw [thin,domain=45:90] plot ({0.5*cos(\x) + 3*cos(135)+3}, {0.5*sin(\x) + 3*sin(135)}) node [above right] {$\psi'$};
    
    \draw [thin,domain=135:180] plot ({cos(\x)+3}, {sin(\x)}) node [above left, xshift=+3pt, yshift=5pt] {$\psi'$};
    
    \coordinate (k) at (3, 0);
    \draw[-] (k) -- node [above right] {$r$} (p');
    
    \draw[dashed] (yaxis |- p') node[left] {$p'_{N_{body}}$} -| (xaxis -| p') node[below] {$p'_{E_{body}}$};

    \fill[blue] (p) circle (3pt) node [below left, color=black] {$p$};
    \fill[orange] (p') circle (3pt) node [above left, color=black] {$p'$};
    \fill[black] (k) circle (2pt) node [below, color=black] {$p_{turn}$};,
\end{tikzpicture}

%% file: tex/results.tex
\section{Simulation results}\label{sec:results}

To demonstrate the functionality of the proposed framework, a set of simulation scenarios are designed to simulate a real world use-case.

\begin{figure}[t]
        \centering
        \includegraphics[width=\columnwidth]{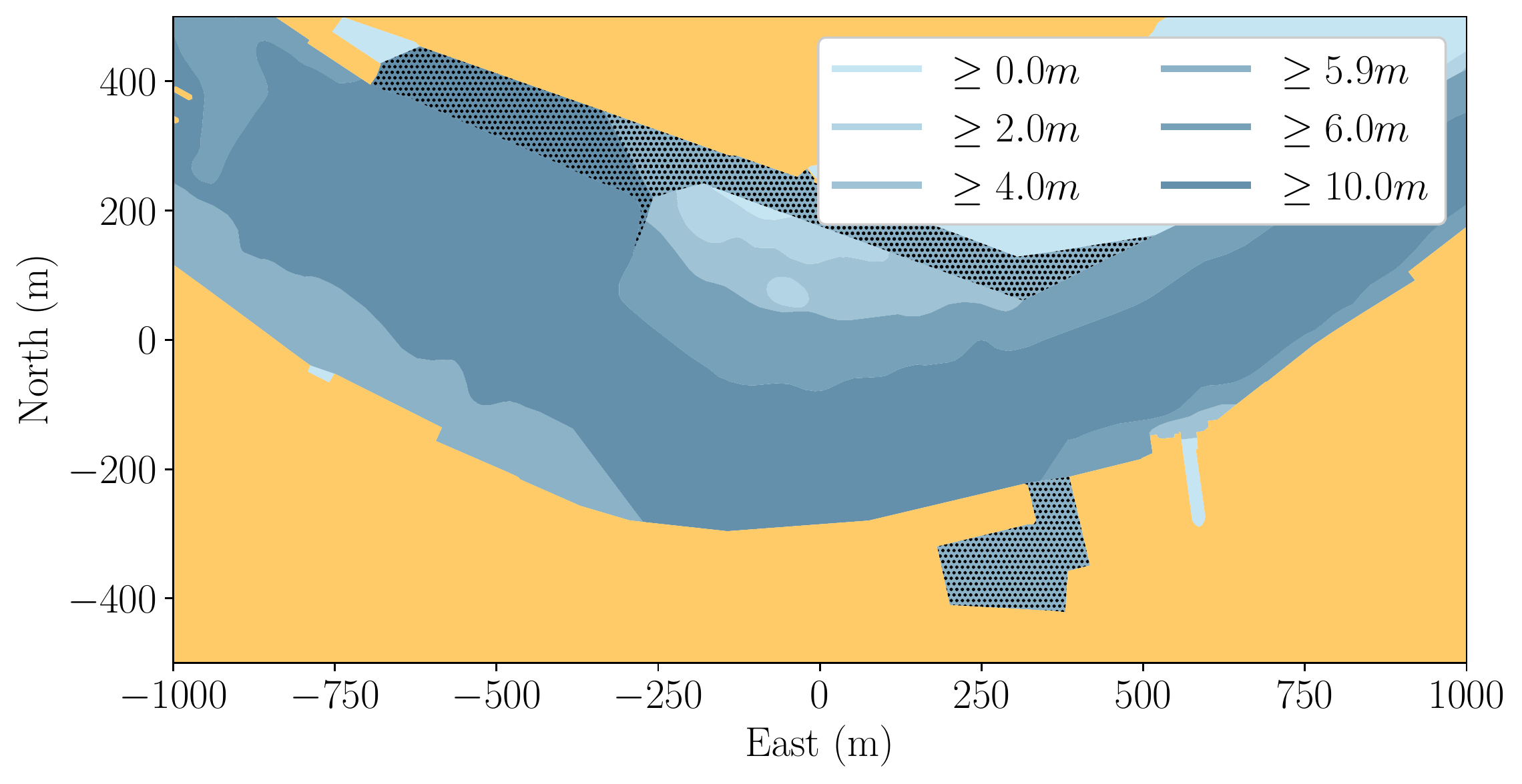}
        \caption{Seachart converted to NED coordinates for the crossing in the Aalborg area.}
        \label{fig:nominal_chart}
\end{figure}

\subsection{Case Study - Greenhopper}

The port of Aalborg, Denmark has commissioned the construction of an 11-meter long passenger ferry, the Greenhopper, that will transport up to 24 passengers between the north and south side of the Aalborg fjord (Limfjorden).
It is the intention that this ferry operates as a fully autonomous system, while being monitored by a remote control center located on shore.

As the Greenhopper will be crossing the fjord, the autonomous system must be able to assess manoeuvrability and apply rule 9 to its decision-making process.
A chart with depth-curves for the Aalborg area is shown in Fig.~\ref{fig:nominal_chart}.

Two scenarios are simulated; one where the target is considered restricted, and one where the target is not restricted.
The simulated ownship will be the Greenhopper ferry, which in both scenarios will be conducting a southbound crossing from the north side to south side of the fjord.
The Greenhopper ferry will have a draught of 0.5 meters and is not considered restricted or draught constrained anywhere in the area.
In both scenarios the target vessel is crossing from port side such that ownship should not give way, and the manoeuvrability restriction is due to the target vessels position and estimated trajectory.
The selection of target vessel dimensions is based on statistical analysis of historical AIS data from Aalborg fjord, where data points are selected defined by the following region
\begin{equation}
    \begin{split}
        \mathcal{R} = \{ (\lambda_i, \phi_i) \mid \, &57.044196 \leq \lambda_i \leq 57.062865 \\ &\land \, 9.909933 \leq \phi_i \leq 9.971545\} .
    \end{split}
\end{equation}
The treatment and preprocessing of the extracted data is performed as described in \cite{enevoldsen2021grounding}. 
The resulting statistical values of length, breadth and draught are shown in Table \ref{tab:ais_stats}.
\begin{table}[tb]
    \centering
    \caption{Statistical overview of vessel dimensions based on historical AIS data}
    \label{tab:ais_stats}
    \begin{tabular}{ l r r r r r} 
        \toprule
        Dimension & mean & median & std & min & max \\
        \midrule
        Length (m) & 50.1 & 33.0 & 39.6 & 5.0 & 240.0 \\
        Breadth (m) & 9.2 & 8.0 & 4.6 & 1.0 & 34.0 \\
        Draught (m) & 3.2 & 2.7 & 1.3 & 0.1 & 9.3 \\
        \bottomrule
    \end{tabular}
\end{table}
The geographic features of the specific area of Limfjorden that the Greenhopper will be operating in, are such that they will impose manoeuvrability restrictions on some of the vessels that frequent the area, simply due to the narrowing of the depth contours as illustrated in Fig.~\ref{fig:nominal_chart}.
From Table \ref{tab:ais_stats} the length and breadth of the target vessel are chosen to reflect the mean value $l=50$ and $b=9.0$.
For the draught a value of $d=5.0$m is chosen, as a vessel with the mean value draught of $\approx 3.0$ m would not be restricted in any region of interest w.r.t. the nominal path, as shown in Fig.~\ref{fig:nominal_chart}.

To trigger the manoeuvrability assessment the target vessel must be violating the comfort zone of ownship.
For these scenarios, the autonomous system onboard ownship is required to maintain a \gls{CPA}$_{req} > 150m$, and the simulations are designed such that \gls{TV} will violate this constraint.
The autonomous system is set to react when $\text{TCPA} = \text{TCPA}_{act} = 3$ minutes.
The reason for this limit is that the autonomous ferry is sailing quite slow ($\approx 2$ knots), and that when a manoeuvre should be performed, the intention should be clear to the other vessels, in other words, it should not give-way before it is clear who it is giving way to.
\begin{table}[tbp]
    \centering
    \caption{Simulation parameters}
    \label{tab:sim_overview}
    \begin{tabular}{ l r c} 
        \toprule
        parameter & value & unit\\
        \midrule
        Simulation time & 7.0 & minutes \\
        Timestep & 1.0 & seconds \\
        Length & 50.0 & $m$ \\
        Breadth & 9.0 & $m$ \\
        Draught & 5.0 & $m$ \\
        Ownship speed & 2.0 & knots \\
        Target speed & 7.0 & knots \\
        $\delta_{swing}$ & 0.2 & $\frac{deg}{m}$\\
        $\alpha$ & 0.4 & -\\
        $\text{TCPA}_{act}$ & 3 & minutes \\
        $\text{CPA}_{req}$ & 150 & $m$ \\
        \bottomrule
    \end{tabular}
\end{table}
A summary of the simulation parameters are shown in Table \ref{tab:sim_overview}.
\begin{figure}[tb]
    \centering
    \begin{subfigure}[b]{0.49\textwidth}
        \centering
        \includegraphics[width=\columnwidth]{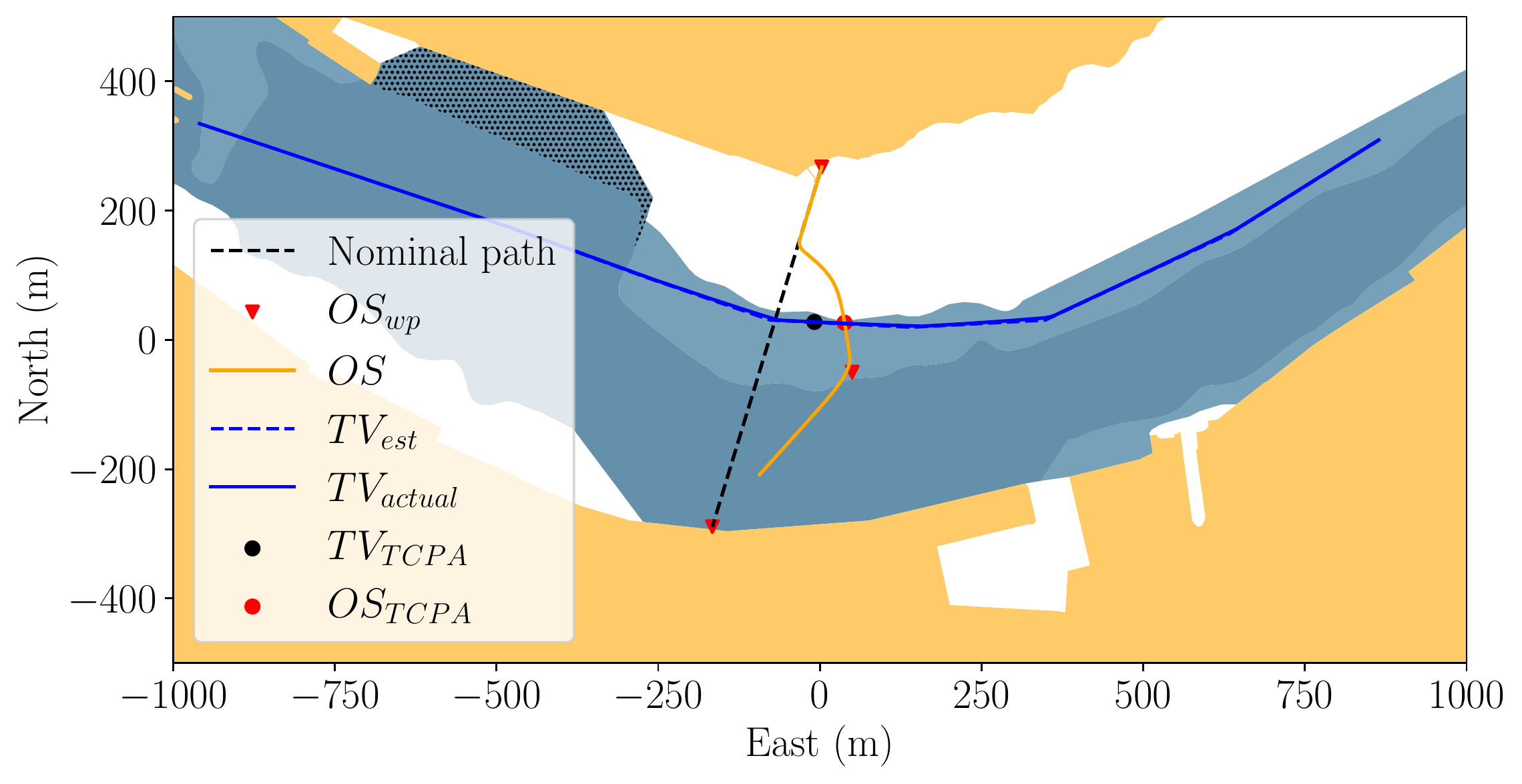}
        \caption{Target is estimated to be manoeuvrability restricted}
        \label{fig:restricted}
    \end{subfigure}
    \hfill
    \begin{subfigure}[b]{0.49\textwidth}
        \centering
        \includegraphics[width=\columnwidth]{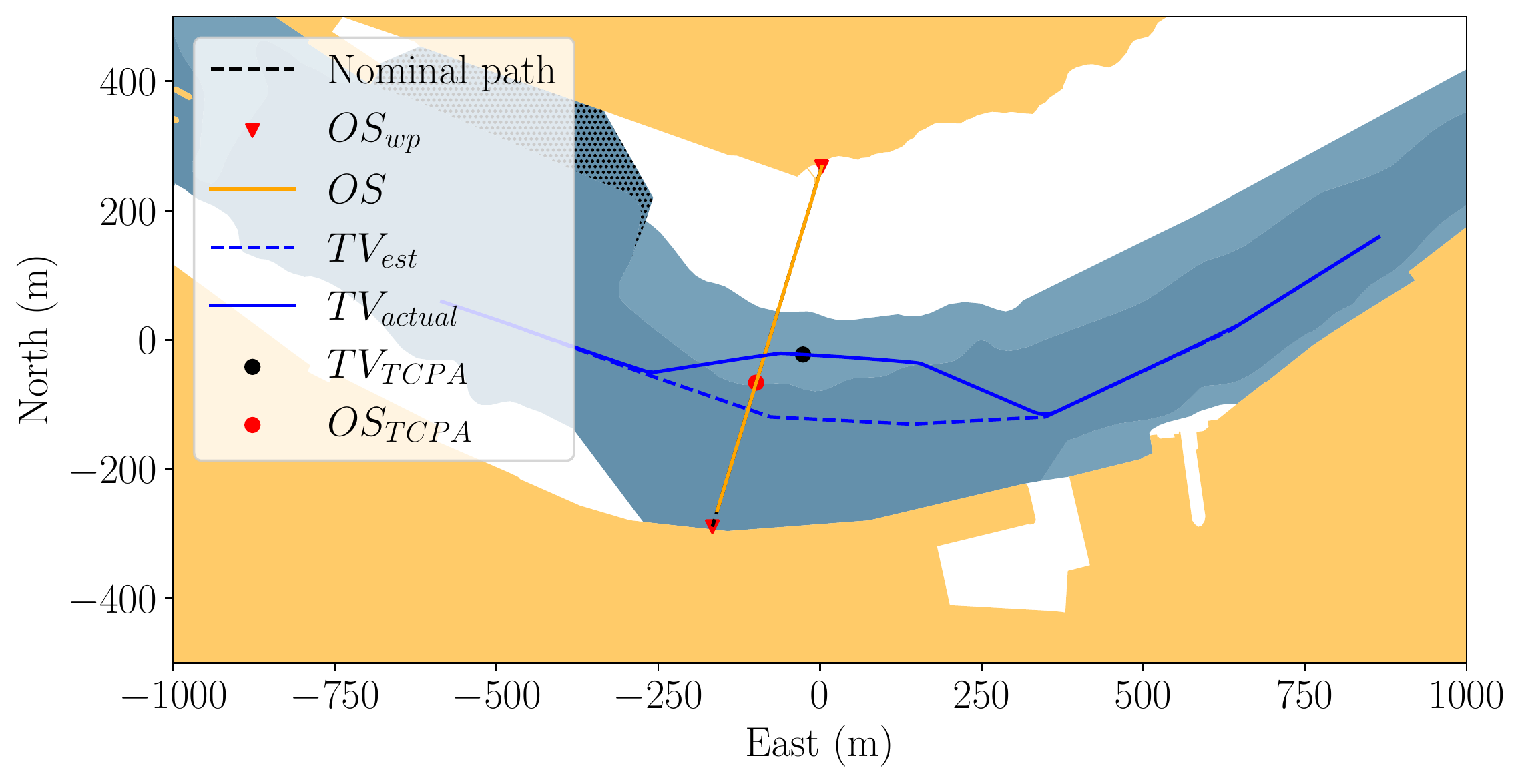}
        \caption{Target is estimated to not be manoeuvrability restricted}
        \label{fig:not_restricted}
    \end{subfigure}
    \caption{Application of the proposed framework on two different scenarios, where a target vessel is crossing from port side.}
    \label{fig:result_figures}
\end{figure}

Figure \ref{fig:restricted} illustrates the simulation scenario where the target vessel has restricted manoeuvrability, and Fig.~\ref{fig:not_restricted} illustrates the scenario where the target is not restricted.
In both scenarios, the position of the ownship and target vessel at \gls{TCPA} are indicated by a red and black dot respectively.
Fig.~\ref{fig:restricted} clearly show that in the scenario where the target is to be considered restricted, the autonomous system will assess the manoeuvrability of the target, and correctly apply rule 9 inverting the port side crossing from a stand-on situation to a give-way situation, as seen from ownship.
Where in Fig.~\ref{fig:not_restricted} the autonomous system will correctly assess that the target vessel has enough open water to manoeuvre, and rule 9 is not applied to the situation.

%% file: tex/conclusion.tex
\section{Conclusion}\label{sec:conclusion}

The paper proposed a framework for evaluating the applicability of \gls{COLREGs} rule 9.
The framework assesses the applicability of rule 9 by performing a manoeuvrability assessment for a single target vessel.
The decision-making logic consists of a single automaton that will trigger the assessment for give-way target vessels.
The assessment is performed by comparing a target to turning characteristics of similar types of vessels, which is then used to evaluate if a target vessel is restricted in terms of its manoeuvrability.
The proposed framework was tested in simulation, in scenarios replicating a real world case-study of an autonomous ferry situated in Aalborg, Denmark.
The paper demonstrated how an autonomous system was able to correctly assess the manoeuvrability of a crossing target vessel, correctly applying the \gls{COLREGs} rule 9 to the situation and reverse the give-way/stand-on relation.

Future work will include extending the framework to incorporate learning-based methods for trajectory estimation for \gls{COLREGs}-compliant decision making.